\def\year{2019}\relax
\documentclass[letterpaper]{article} 
\usepackage{aaai19}  
\usepackage{times}  
\usepackage{helvet}  
\usepackage{courier}  
\usepackage{url}  
\usepackage{graphicx}  
\usepackage{amsmath}
\usepackage{amssymb}
\usepackage{booktabs}
\usepackage{multirow}
\usepackage{makecell}
\usepackage{soul}
\usepackage{appendix}

\newcommand{\citet}[1]
{\citeauthor{#1} \shortcite{#1}}
\newcommand{\citep}{\cite}

\renewcommand{\vector}[1]{\mathbf{#1}}
\renewcommand{\matrix}[1]{\mathbf{#1}}

\usepackage{color}

\pdfoutput=1

\begin{document}
%
\title{Combining Fact Extraction and Verification with \\ Neural Semantic Matching Networks}
\author{Yixin Nie \;\;\;\; Haonan Chen \;\;\;\; Mohit Bansal\\
Department of Computer Science\\
University of North Carolina at Chapel Hill\\
\texttt{\{yixin1, haonanchen, mbansal\}@cs.unc.edu}\\
}
\maketitle
\begin{abstract}
The increasing concern with misinformation has stimulated research efforts on automatic fact checking. The recently-released FEVER dataset introduced a benchmark fact-verification task in which a system is asked to verify a claim using evidential sentences from Wikipedia documents.
In this paper, we present a connected system consisting of three homogeneous neural semantic matching models that conduct document retrieval, sentence selection, and claim verification jointly for fact extraction and verification.
For evidence retrieval (document retrieval and sentence selection), unlike traditional vector space IR models in which queries and sources are matched in some pre-designed term vector space, we develop neural models to perform deep semantic matching from raw textual input, assuming no intermediate term representation and no access to structured external knowledge bases. We also show that Pageview frequency can also help improve the performance of evidence retrieval results, that later can be matched by using our neural semantic matching network.
For claim verification, unlike previous approaches that simply feed upstream retrieved evidence and the claim to a natural language inference (NLI) model, we further enhance the NLI model by providing it with internal semantic relatedness scores (hence integrating it with the evidence retrieval modules) and ontological WordNet features.
Experiments on the FEVER dataset indicate that (1) our neural semantic matching method outperforms popular TF-IDF and encoder models, by significant margins on all evidence retrieval metrics, (2) the additional relatedness score and WordNet features improve the NLI model via better semantic awareness,
and (3) by formalizing all three subtasks as a similar semantic matching problem and improving on all three stages, the complete model is able to achieve the state-of-the-art results on the FEVER test set (two times greater than baseline results).\footnote{Code: {\scriptsize\url{https://github.com/easonnie/combine-FEVER-NSMN}}}

\end{abstract}

\section{Introduction}
\label{sec:introduction}

The explosion of online textual content with unknown integrity and verification raises an important concern about misinformation such as fake news, socio-political deception, and online rumors. 
This problem of misinformation could potentially produce uncontrollable and harmful social impacts, thus stimulating recent research efforts on leveraging modern machine learning techniques for automatic fact checking.
The recent release of the Fact Extraction and VERification~\cite{thorne2018fever} (FEVER) dataset not only provides valuable fuel for applying data-driven neural approaches on evidence retrieval and claim verification, but also introduces a standardized, benchmark task of the automatic fact checking. 
In this FEVER shared task, a system is asked to verify an input claim with potential evidence in about 5 million Wikipedia documents, and label it as ``\textsc{Supports}", ``\textsc{Refutes}", or ``\textsc{Not Enough Info}" if the evidence can support, refute, or not be found for the claim, respectively. Fig. \ref{fig:example} shows an example of the task.
The task is difficult in two aspects. First, accurate selection of potential evidence from a huge knowledge base, w.r.t. an arbitrary claim requires a thoughtful system design and results in a trade-off between retrieval performance and computational resources. Moreover, even with ground truth evidence, the verification sub-task of predicting the relation between evidence and the claim is still a long-existing open problem.\footnote{The task is often termed as natural language inference (NLI).}

\begin{figure}[t]
\fbox{
\begin{minipage}{0.95\linewidth}
\begin{small}
\textbf{Claim:} Giada at Home was only available on DVD.\\

\textbf{[{\ttfamily wiki/Giada\_at\_Home}]}\\
Giada at Home is a television show hosted by Giada De Laurentiis. \ul{It first aired on October 18, 2008 on the Food Network}.\\

\textbf{[{\ttfamily wiki/Food\_Network}]}\\
\ul{Food Network is an American basic cable and satellite television channel} that is owned by Television Food Network, G.P., a joint venture and general partnership between Discovery, Inc. (which owns 70\% of the network) and Tribune Media (which owns the remaining 30\%).\\

\textbf{Label:} Refutes
\end{small}
\end{minipage}}

\caption{Example of FEVER task. Given the claim, the system is required to find evidential sentences in the entire Wikipedia corpus and label it as ``\textsc{Supports}", ``\textsc{Refutes}", or ``\textsc{Not Enough Info}" \cite{thorne2018fever}.
}
\label{fig:example}
\end{figure}

In this work, we propose a joint system consisting of three connected homogeneous networks for the 3-stage FEVER task of document retrieval, sentence selection, and claim verification and frame them as a similar semantic matching problem.
In the document retrieval phase, the corresponding sub-module selects documents from the entire Wikipedia corpus by keyword matching and uses a neural semantic matching network for further document ranking. In the sentence selection phase, we use the same neural architecture trained with an annealed sampling method to select evidential sentences by conducting semantic matching between each sentence from retrieved pages and the claim. Finally, we build a neural claim verifier by integrating upstream semantic relatedness features (from the sentence selector) and injecting ontological knowledge from WordNet into a similar neural semantic matching network for natural language inference (NLI), and train it to infer whether the retrieved evidence supports or refutes the claim, or state that the evidence is not enough to decide the correctness of the claim.

Overall, our unified neural-semantic-matching model for fact extraction and verification, which includes a three-fold contribution: (1) Unlike traditional IR methods e.g., TF-IDF, in which queries and sources are matched in some vector space according to pre-designed terms and pre-calculated weightings, we explore the possibility of using a neural semantic matching network for evidence retrieval and show that by assuming no intermediate term representation, neural networks can learn their own optimal representation for semantic matching at the granularity of sentences and significantly outperform term-weighting based methods. We also show that external Pageview frequency information can provide comparable and complementary  discriminative information w.r.t. the neural semantic matching network for document ranking. (2) In contrast to previous work, in which upstream retrieved evidence are simply provided to downstream NLI models, we combined the evidence retrieval module with the claim verification module, by adding semantic relatedness scores to the NLI models, and further improve verification performance by using additional semantic ontological features from WordNet. (3) Rather than depending on structured machine-friendly knowledge bases such as Freebase \cite{freebase} and DBpedia \cite{dbpedia}, we formalize the three subtasks as a similar textual semantic matching problem and propose one of the first neural systems that are able to conduct evidence retrieval and fact verification using raw textual claims and sentences directly from Wikipedia as input, and achieves the state-of-the-art results on the FEVER dataset, which could serve as a new neural baseline method for future advances on large-scale fact checking.

\section{Related Works}
\label{sec:related_works}
\noindent\textbf{Open Domain Question Answering}: Recent deep learning-based open domain question-answering (QA) systems follow a two-stage process including document retrieval, selecting potentially relevant documents from a large corpus (such as Wikipedia), and reading comprehension, extracting answers (usually a span of text) from the selected documents. \citet{chen2017readingatscale} was the first  to successfully applied this framework to open domain QA, obtaining state-of-the-art results on several QA benchmarks \cite{rajpurkar2016squad,baudivs2015modeling,miller2016key,berant2013semantic}. Following their work, \citet{dhingra2017quasar} introduced new benchmarks, \citet{wang2017r} extended the framework by adding feedback signals from comprehension to the upstream document retriever, and \citet{kratzwald2018adaptive} proposed to adaptively adjust the number of retrieved documents. FEVER shares the similar retrieval problem as open domain QA, while the end task is claim verification.

\noindent
\textbf{Information Retrieval}:
Recent success in deep neural networks has brought increasing interest in their application to information retrieval (IR) tasks~\cite{huang2013learning,guo2016deep,mitra2017learning,dehghani2017neural}.
Although IR tasks also look at sentence-similarity, as discussed in \citet{guo2016deep}, they are more about relevance-matching, in which the match of specific terms plays an important role.
As the end goal of FEVER is verification, its retrieval aspect is more about sentences having the same semantic meaning, and therefore, we approach the problem via natural language inference techniques, instead of relevance-focused IR methods.

\noindent\textbf{Natural Language Inference}: NLI is a task in which a system is asked to classify the relationship between a pair of premise and hypothesis as either entailment, contradiction or neutral.
Large annotated datasets such as the Stanford Natural Language Inference \cite{snli:emnlp2015} (SNLI) and the Multi-Genre Natural Language Inference \cite{williams2017broad} (Multi-NLI) have promoted the development of many different neural NLI models \cite{RSE,conneau2017maxout_encoder,decomposable,chen2016esim,DIIN,DRLSTM} that achieve promising performance. The task of NLI is framed as a typical semantic matching problem which exists in almost all kinds of NLP tasks. Therefore, in addition to the final claim verification subtask, we also formalize the other two FEVER subtasks of document retrieval and sentence selection as a similar problem and solve them using a homogeneous network.

\noindent\textbf{Automatic Fact Checking Datasets}: Besides FEVER, there are a number of fact checking datasets and tasks. \citet{vlachos2014factchecking} collected and released 211 labeled claims in the political domain with evidence hyperlinks. \citet{wang2017liar} provided 12.8K labeled claims with some meta-data but without structured evidence. The 2017 Fake News Challenge \cite{FNC} released a dataset consisting of approximately 50K headline and body pairs derived from the dataset in \citet{ferreira2016emergent}. A retrospective analysis of the models for this challenge is provided in \citet{fakenews_retro_analysis}. The Triple Scoring Task at the WSDM Cup 2017 \cite{triple_score_task@2017} requires a system to verify a relation triple based on 818,023 triples from two Freebase relations. \citet{perez2017automatic} introduces two novel datasets for the task of fake news detection, covering seven different news domains. In this work, we experiment on FEVER, which is the first task that requires and measures models' joint ability for both evidence retrieval and verification, and provides a benchmark evaluation.

\noindent\textbf{Other FEVER Systems:} There are other proposed methods during the FEVER Shared Task in which our system, \citet{yoneda2018ucl_2nd} and \citet{hanselowski2018ukp_3rd} are the top three on the leaderboard. The most apparent differences between our system and other methods are: (1) our method uses a homogeneous semantic matching network to tackle all the subtasks while others utilize different models for different subtasks and (2) our vNSMN verification model takes the concatenation of all the retrieved sentences (together with their relatedness score) as input, while other systems use existing NLI models to compare each of the evidential sentences with the claim and then apply another aggregation module to merge all the outputs for final prediction.

\section{FEVER: Fact Extraction and VERification}
\label{sec:fact_extraction_and_claim_verification}

\begin{figure*}[t]
\centering
\includegraphics[width=0.95\textwidth]{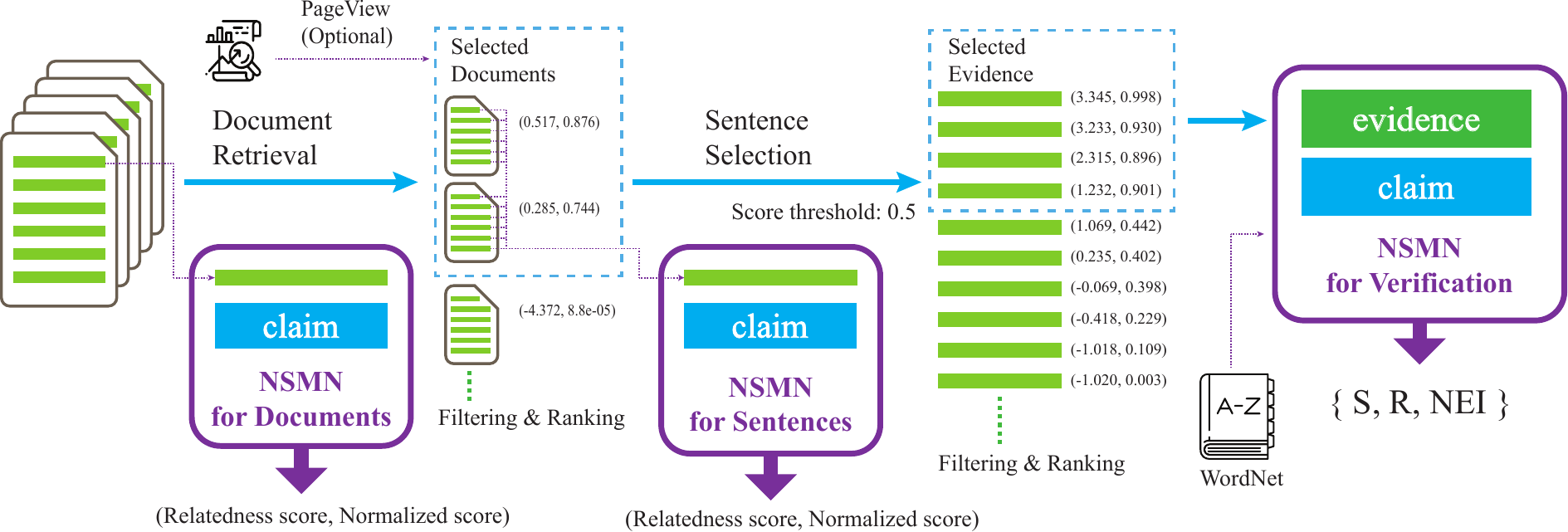}
\caption{System Overview: Document Retrieval, Sentence Selection, and Claim Verification.}
\label{fig:fig_system_overview}
\end{figure*}

\subsection{Task Formalization}
FEVER~\cite{thorne2018fever} is a comprehensive task in which a system is asked to verify an arbitrary claim with potential evidence extracted from a huge list of Wikipedia documents, or states that the claim is non-verifiable it cannot find enough evidence.
Suppose $P_{i} \in \mathbb{P}$ denotes an individual Wikipedia document and $\mathbb{P}=\{P_{0}, P_{1}, ...\}$ is the set of all the provided documents. $P_{i}$ is also an array of sentences, namely $P_{i}=\{s^0_{i},s^1_{i},s^2_{i},...s^m_{i}\}$ with each $s^j_{i}$ denoting the $j$-th sentence in the $i$-th Wikipedia document (where $s^0_{i}$ is the title of the document).
The inputs of each example are a textual claim $c_i$ and $\bigcup P_{i}$, the union of all the sentences in each provided Wikipedia document. The output should be a tuple $(\hat{E_i}, \hat{y_i})$ where $\hat{E_i}=\{s^{e_0}, s^{e_1},...\} \subset \bigcup P_{i}$, representing the set of evidential sentences for the given claim, and $\hat{y_i} \in \{\textit{S},\textit{R}, \textit{NEI}\}$\footnote{$\textit{S},\textit{R}, \textit{NEI}$ represent ``\textsc{Supports}'', ``\textsc{Refutes}'' and ``\textsc{Not Enough Info}'', respectively.}, the predicted label for the claim. Suppose the ground truth evidence and label are $E_i$ and $y_i$. For a successful verification of a given claim $c_i$, the system should produce a prediction tuple $(\hat{E_i}, \hat{y_i})$ satisfying ${E_i} \subseteq \hat{E_i}$ and ${y_i} = \hat{y_i}$.
The dataset details (size and splits) are discussed in~\citet{thorne2018fever}.
As described above, the single prediction is considered to be correct if and only if both the label is correct and the predicted evidence set (containing at most five sentences\footnote{This constraint is imposed in the FEVER Shared Task because in the blind test set, all claims can be sufficiently verified with at most 5 sentences of evidence.}) covers the annotated evidence set. This score is named as \textbf{FEVER Score}.

\section{Our Model}
\label{sec:the_system}
In this section, we first describe the architecture of our Neural Semantic Matching Network (NSMN), and then elaborate on the three subtasks of document retrieval, sentence selection and claim verification, especially how these different sub-tasks can be treated as a similar semantic matching problem and be consistently resolved via the homogeneous NSMN architecture. See Fig. \ref{fig:fig_system_overview} for our system's overview.

\subsection{Neural Semantic Matching Network}
\label{sec:nsmn}
The Neural Semantic Matching Network (NSMN) is the key component in each of our sub-modules that performs semantic matching between two textual sequences.
Specifically, the NSMN contains four layers as described below.\footnote{Our NSMN model is a modification of the Enhanced Sequential Inference Model (ESIM)~\cite{chen2016esim}, where we add shortcut connections from input to matching layer and change  output layer to only max-pool plus one affine layer with rectifier activation. These modifications are based on validation results.}

\noindent\textbf{Encoding Layer}: Suppose the two input sequences are $\matrix{U} \in \mathbb{R}^{d_0 \times n}$ and $\matrix{V} \in \mathbb{R}^{d_0 \times m}$, the encoding layer is one bidirectional LSTM that encodes each input token with its contexts:
\begin{align}
\matrix{\bar U}&=\text{BiLSTM}(\matrix{U})\in\mathbb{R}^{d_1 \times n},\\
\matrix{\bar V}&=\text{BiLSTM}(\matrix{V})\in\mathbb{R}^{d_1 \times m},
\end{align}
where $d_0$ and $d_1$ are input and output dimensions of the encoding layer and $n$ and $m$ are lengths of the two sequences.

\noindent\textbf{Alignment Layer}: This produces an alignment between the two input sequences based on the encoding of tokens computed above. The alignment matrix is computed as:
\begin{align}
    \matrix{E} = \matrix{\bar U}^{\top}  \matrix{\bar V} \in \mathbb{R}^{n \times m}.
\end{align}
Each element $e_{ij}$ in the matrix indicates the alignment score between $i$-th token in $\matrix{U}$ and $j$-th token in $\matrix{V}$. Then, for each input token, the model computes the relevant semantic content from the other sequence using the weighted sum of encoded tokens according to the normalized alignment score:
\begin{align}
\matrix{\tilde U} &= \matrix{\bar V} \cdot \text{Softmax}_\mathrm{col}(\matrix{E^\top}) \in \mathbb{R}^{d_1 \times n},\\
\matrix{\tilde V} &= \matrix{\bar U} \cdot \text{Softmax}_\mathrm{col}(\matrix{E}) \in \mathbb{R}^{d_1 \times m},
\end{align}
where $\text{Softmax}_\mathrm{col}$ denotes column-wise softmax function. $\matrix{\tilde U}$ is the aligned representation from $\matrix{\bar V}$ to $\matrix{\bar U}$ and vice versa for $\matrix{\tilde V}$. The aligned and encoded representations are combined:
\begin{align}
\matrix{S} &= f([\matrix{\bar U}, \matrix{\tilde U}, \matrix{\bar U} - \matrix{\tilde U}, \matrix{\bar U} \circ \matrix{\tilde U}]) \in \mathbb{R}^{d_2 \times n}, \\
\matrix{T} &= f([\matrix{\bar V}, \matrix{\tilde V}, \matrix{\bar V} - \matrix{\tilde V}, \matrix{\bar V} \circ \matrix{\tilde V}]) \in \mathbb{R}^{d_2 \times m},
\end{align}
where $f$ is one affine layer with a rectifier as an activation function and $\circ$ indicates element-wise multiplication.

\noindent\textbf{Matching Layer}:
The matching layer takes the upstream compound aligned representation and performs semantic matching between two sequences via a recurrent network as:
\begin{align}
\matrix{P}&=\text{BiLSTM}([\matrix{S}, \matrix{U^*}])\in\mathbb{R}^{d_3 \times n},\\
\matrix{Q}&=\text{BiLSTM}([\matrix{T},\matrix{V^*}])\in\mathbb{R}^{d_3 \times m}.
\end{align}
Note that $\matrix{U^*}$ and $\matrix{V^*}$ are additional input vectors for each token provided to the matching layer via a shortcut connection. In this work, $\matrix{U^*}$ and $\matrix{V^*}$ are sub-channels of the input $\matrix{U}$ and $\matrix{V}$ without GloVe, aimed to facilitate the training.

\noindent\textbf{Output Layer}:
The two matching sequences are projected onto two compressed vectors by max-pooling along the row axes. The vectors, together with their absolute difference and element-wise multiplication, are mapped to the final output $\vector{m}$ by a function $h$.
\begin{align}
&\vector{p} = \text{Maxpool}_\mathrm{row}(\matrix{P})\in\mathbb{R}^{d_3},\\
&\vector{q} = \text{Maxpool}_\mathrm{row}(\matrix{Q})\in\mathbb{R}^{d_3},\\
& h (\vector{p},\vector{q}, |\vector{p} - \vector{q}|, \vector{p} \circ \vector{q}) = \vector{m},
\end{align}
where function $h$ denotes two affine layers with a rectifier being applied on the output of the first layer.

The final output vector is different for the extraction versus the verification subtasks. For the extraction subtasks (document retrieval and sentence selection), $\vector{m} = \left< m^+, m^- \right>$, where $m^+\in \mathbb{R}$ is a scalar value indicating the score for selecting the current sentence as evidence, and $m^-$ gives the score for discarding it.
For claim verification, $\vector{m} = \left< m_s, m_r, m_n\right>$, where the elements of the vector denote the score for predicting the three labels, namely \textsc{Supported}, \textsc{Refute}, and \textsc{NEI}, respectively.

\subsection{Three-Phase Procedure}

\paragraph{1. Document Retrieval}
Document retrieval is the selection of Wikipedia documents related to a given claim. This sub-module handles the task as the following function:
\begin{equation}
f(c_i, \mathbb{P}) = D_{c_i},
\end{equation}
where a claim $c_i$ and the complete collection of documents $\mathbb{P}$ are mapped to a set of indices $D_{c_i}$ such that $\{P_i \mid i\in D_{c_i}\}$ is the set of documents required to verify the claim. The task can be viewed as selecting a fine-grained document subset from a universe of documents by comparing each of them with the input claim. 
We observe that 10\% of the documents have disambiguation information in their titles, such as ``Savages (band)''; the correct retrieval of these documents requires semantic understanding and we will rely on the NSMN to handle this problem. For clarity, we will call these documents as ``disambiguative" documents for later use.

Because the number of documents in the collection are huge, conducting semantic matching between all the documents from the collection with the claim is computationally intractable. Hence, we start the retrieval by applying a \textbf{keyword matching} step to narrow down the search space.\footnote{On average, keyword matching returns 8 pages for each claim.} The details about the keyword matching is described in the supplementary.
Next, we give all the documents that are not ``disambiguative'' the highest priority score and rank the ``disambiguative'' documents by comparing each of them with the claim using the NSMN (Sec.~\ref{sec:nsmn}). The document is represented as the concatenation of its title and the first sentence. The matching score between the claim $c_i$ and the $j$-th document is computed as:
\begin{align}
    \left< m^+, m^- \right> &= \text{NSMN}(c_i, [t_j, s^0_j]), \\
p(x=1 \mid c_i, j) &= \frac{e^{m^+}}{e^{m^+}+e^{m^-}},
\end{align}
where $x \in \{0, 1\}$\footnote{$1$ indicates to choose the document and $0$ otherwise} indicates whether to choose the $j$-th document, $m^+$ is the score for prioritizing the document, and $p(x=1 \mid c_i, j) \in (0, 1)$ is the normalized score for $m^+$. Note that $m^+$ can also be viewed as the global semantic relatedness of a document to a claim. A document having a higher $m^+$ value than others, w.r.t. a claim, indicates that it is semantically more related to the claim.

To summarize, document retrieval phase involves steps:
\begin{itemize}
    \item Building a candidate subset with keyword matching on all documents in the collections;
    \item Adding all the documents that are not ``disambiguative'' to the resulting list;\footnote{If there are multiple ``disambiguative'' documents, we randomly select at most five documents.}
    \item Calculating the $p(x=1 \mid c_i, j)$ and $m^+$ value for ``disambiguative'' documents in the candidate set using NSMN;
    \item Filtering out the documents having $p(x=1 \mid c_i, j)$ lower than some threshold $P^{d}_\mathrm{th}$;
    \item Sorting the remaining documents by their $m^+$ values and adding the top $k$ documents to the resulting list.
\end{itemize}

\paragraph{2. Sentence Selection}
Sentence selection is the extraction of evidential sentences from the retrieved documents regarding a claim, formalized as the following function:
\begin{equation}
g(c_i, \bigcup_{i \in D_{c_i}} P_{i})=E_{c_i},
\end{equation}
which takes a claim and the union of the sentence set of each retrieved document as inputs and outputs a subset of sentences $E_{c_i} \subseteq \bigcup_{i \in D_{c_i}} P_{i}$ as the evidence set. Similar to document retrieval, sentence selection can also be treated as conducting semantic matching between each sentence $s_j \in \bigcup_{i \in D_{c_i}}$ and the claim $c_i$ to select the most plausible evidence set. Since the search space is already narrowed down to a controllable size by the document retrieval, we can directly traverse all the sentences and compare them with the claim using NSMN.
The selection is done via these steps:
\begin{itemize}
    \item Calculating the $p(x=1 \mid c_i, j)$ and $m^+$ value for all the sentences in the retrieved documents;
    \item Filtering out the sentences having $p(x=1 \mid c_i, j)$ lower than some threshold $P^s_\mathrm{th}$;
    \item Sorting sentences by their $m^+$ values and adding the top 5 sentences to the resulting list.
\end{itemize}

\paragraph{3. Claim Verification} 
This final sub-task requires logical inference from evidence to the claim, which is defined as:
\begin{equation}
    h(E_{c_i}, c_i) = y,
\end{equation}
where $E_{c_i}$ is the set of evidential sentences and $y \in \{\textit{S},\textit{R}, \textit{NEI}\}$ is the output label.

We use the similar neural semantic matching network with additional token-level features for the final claim verification (similar to NLI task). The input premise is the concatenation of all sentences from the upstream evidence set, and the input hypothesis is the claim. More importantly, besides GloVe and ELMo\footnote{We used GloVe+ELMo because their combination gives a comprehensive+contextualized lexical representation of the inputs.} embeddings, the concatenation of the following three additional token-level features are added as task-specific token representations to further improve the verification accuracy:

\noindent
\textbf{WordNet:} 30-dimension indicator features regarding ontological information from Wordnet. The 30 dimensions are divided into 10 embedding channels corresponding to 10 hypernymy/antonymy and edge-distance based phenomena, as shown in Table~\ref{tab:wordnet_feature}. For the current token, if one of these 10 phenomena is true for any word in the other sequence, that phenomenon's indicator feature will be fired. As we also want to differentiate whether the token is in the evidence or the claim, we use 3 elements for each channel with first two elements indicating the position and the last element for the feature indication. For example, if a token in the evidence fires then the vector will be [1, 0, 1] and if the token is in the claim it will be [0, 1, 1].

\noindent
{\bf Number:} We use 5-dimension real-value embeddings to encode any unique number token. This feature assists the model in identifying and differentiating numbers.

\noindent
{\bf Normalized Semantic Relatedness Score:} Two normalized relatedness scores, namely the two $p(x=1 \mid c_i, j)$ values produced by the document and sentence NSMN, respectively. These scores are served as 2-dimension features for each token of the evidence. We add these features to connect the three subtask modules strongly, i.e., help the claim verification model better focus on the evidence, based on the semantic relatedness strength between the current evidence and the claim.

\begin{table}[t]
\centering
\begin{small}
\begin{tabular}{l}
\toprule
Exact same lemma\\
Antonym\\
Hyponym\\
Hypernym\\
Hyponym with 1-edge distance in WN topological graph\\
Hypernym with 1-edge distance in WN topological graph\\
Hyponym with 2-edge distance in WN topological graph\\
Hypernym with 2-edge distance in WN topological graph\\
Hyponym with distance $>$ 2 edges in WN topological graph\\
Hypernym with distance $>$ 2 edges in WN topological graph\\
\bottomrule
\end{tabular}
\end{small}
\caption{10 indicator features in WordNet embedding.}
\label{tab:wordnet_feature}
\end{table}

\noindent
{\bf Evidence Enhancement (Optional):} An optional step that augments the current evidence set. By default, we apply evidence enhancement before evaluation. Details are provided in the supplementary.

\section{Implementation and Training Details}

\begin{table*}[t]
\centering
\begin{small}
\begin{tabular}{ccccccccc}
\toprule
\multirow{2}{*}{\textbf{Model}} &
\multicolumn{4}{c}{Entire Dev Set} &
\multicolumn{4}{c}{Difficult Subset ($>$10\%)} \\
\cmidrule(lr){2-5}
\cmidrule(lr){6-9}
 & \textbf{OFEVER} & Acc. & Recall & F1 & \textbf{OFEVER} & Acc. & Recall & F1\\
\midrule
FEVER Baseline & 70.20 & -- & -- & -- & -- & -- & -- & --\\
KM & 88.86 & 44.90 & 83.30 & 58.35 & 60.15 & 23.89 & 60.15 & 34.20 \\
KM + Pageview  & 91.98 & 45.90 & 87.98 & 60.32 & 85.61 & 29.32 & 85.61 & 43.68 \\
KM + TF-IDF  & 91.63 & 42.83 & 87.45 & 57.50 & 85.60 & 28.66 & 85.60 & 42.94 \\
KM + dNSMN & 92.34 & 52.70 & 88.51 & 66.06 & 87.93 & 31.71 & 87.93 & 46.61 \\
KM + Pageview + dNSMN & \textbf{92.42} & \textbf{52.73} & \textbf{88.63} & \textbf{66.12} & \textbf{88.73} & \textbf{31.90} & \textbf{88.72} & \textbf{46.93} \\
\multicolumn{9}{l}{\footnotesize{\textbf{\textit{k = 5}}}}\\
\midrule
FEVER Baseline & 77.24 & -- & -- & -- & -- & -- & -- & --\\
KM & 90.69 & 42.61 & 86.04 & 56.99 & 74.34 & 23.19 & 74.34 & 35.36 \\
KM +  Pageview & 92.69 & 42.92 & 89.04 & 57.92 & 90.52 & 24.89 & 90.52 & 39.05 \\
KM + TF-IDF & 92.38 & 39.57 & 88.57 & 54.70 & 89.88 & 23.94 & 89.88 & 37.80 \\
KM + dNSMN & \textbf{92.82} & 51.04 & \textbf{89.23} & \textbf{64.94} & 91.33 & 28.30 & 91.33 & 43.21 \\
KM + Pageview + dNSMN & 92.75 & \textbf{51.06} & 89.13 & 64.93 & \textbf{91.36} & \textbf{28.38} & \textbf{91.37} & \textbf{43.30} \\
\multicolumn{9}{l}{\footnotesize{\textbf{\textit{k = 10}}}}\\
\bottomrule
\end{tabular}
\end{small}
\caption{Performance of different document retrieval methods. \textbf{\textit{k}} indicates the number of retrieved documents. The last four columns show results on the difficult subset that includes more than 10\% of dev set. dNSMN = document retrieval Neural Semantic Matching Network. `KM'=Keyword Matching.}
\label{tab:doc_retrieval_ablation}
\end{table*}

\noindent\textbf{Document Retrieval}: The neural semantic matching network is trained by optimizing cross-entropy loss using the ``disambiguative'' documents containing ground truth evidence as positive examples and all other ``disambiguative'' document as negative examples. We used Adam optimizer \cite{kingma2014adam} with a batch size of 128. We also consider using Pageview frequency resources, TF-IDF method after keyword matching for re-ranking the documents and compare their results in the experiments.

\noindent\textbf{Sentence Selection}: We trained neural sentence selector using the FEVER training set by optimizing cross-entropy loss with ground truth evidence as positive examples and all other sentences in the candidate pool from the document retriever as negative examples. We used Adam optimizer \cite{kingma2014adam} with a batch size of 128.
We applied an {\bf annealed sampling} strategy to gradually increase the portion of positive examples after every epoch. Concretely, each negative example in the training data will be added to the next training epoch with a decreasing probability $p_e$. $p_e$ starts from $0.5$ at the first epoch and decreases $0.1$ after each epoch and is reset to $0.02$ when $p_e \le 0$. The intuition behind annealed sampling is that we want the model to be more tolerant about selecting sentences while being discriminative enough to filter out apparent negative sentences. We also experiment with using TF-IDF method and a Max-Pool sentence encoder that gives a comprehensive representation for sentence modeling \cite{conneau2017maxout_encoder} in place of the NSMN for sentence selection.

\noindent\textbf{Claim Verification}: We trained our verification NSMN using ground truth labels in FEVER training set. For verifiable claims, the input evidence is the provided ground truth supporting or refuting evidence. For non-verifiable claims with no given evidence, we randomly sample 3-5 sentences from candidate sentence pool with equal probability given by the upstream selected sentence. We use Adam optimizer for training the model with a batch size of 32.

\section{Results and Analysis}
\label{sec:results_and_analysis}
\begin{table*}[t]
\centering
\begin{small}
\begin{tabular}{lcccccccc}
\toprule
\multirow{2}{*}{\textbf{Method}} & \multicolumn{4}{c}{Entire Dev Set} &
\multicolumn{4}{c}{Difficult Subset ($>$12\%)} \\ 
\cmidrule(lr){2-5}
\cmidrule(lr){6-9}
& \bf{OFEVER} & Acc. & Recall & F1
& \bf{OFEVER} & Acc. & Recall & F1\\
\midrule
FEVER Baseline & 62.81 & -- & -- & -- & -- & -- & -- & -- \\
TF-IDF & 83.77 & 34.16 & 75.65 & 47.07 & 53.01 & 38.54 & 51.01 & 44.63\\
Max-Pool Enc. & 84.08 & 59.52 & 76.13 & 66.81 & 73.68 & 54.13 & 73.68 & 62.41\\
sNSMN w/o AS & 86.65 & \bf{69.43} & 79.98 & \bf{74.33} & 68.34 & \bf{67.82} & 68.34 & \bf{68.08}\\
sNSMN w. AS & \bf{91.19} & 36.49 & \bf{86.79} & 51.38 & \bf{81.44} & 34.56 & \bf{81.44} & 48.53\\
\bottomrule
\end{tabular}
\end{small}
\caption{Different methods for sentence selection on dev set. `Enc.'= Sentence Encoder. `AS'= Annealed Sampling. The {\bf OFEVER} column shows Oracle FEVER Score. The other three columns show the evidence accuracy, recall, and F1.}
\label{tab:sent_selection_ablation}
\end{table*}

In this section, we present extensive ablation studies for each module in our system, and report our final full-system results.
When evaluating the performance of document retrieval and sentence selection, we compare the upper bound of the FEVER score (or oracle score \textbf{OFEVER}) by assuming perfect downstream systems.\footnote{\textbf{OFEVER} is the same metric as the ``Oracle Accuracy'' in the original baseline in~\cite{thorne2018fever}.}
Besides that, we also provide other metrics (i.e., F1 and label accuracy) for analyzing different submodules. For simplicity, we name dNSMN for document retrieval NSMN, sNSMN for sentence selection NSMN, and vNSMN for verification NSMN.

\paragraph{Document Retrieval Results}
In Table \ref{tab:doc_retrieval_ablation}, we compare the performance of different methods for document retrieval on the entire dev set and on a difficult subset of the dev set. This subset is built by choosing examples having at least one evidence contained in the ``disambiguative'' document. We hypothesize that the correct retrieval of these documents will be more semantically demanding and challenging. To begin with, the keyword matching method (getting 88.86\% and 90.69\% oracle score for $k=5$ and $10$) is better than the FEVER baseline (getting 70.20\% and 77.24\% oracle score for $k=5$ and $10$) with TF-IDF in \citet{chen2017readingatscale}. This is due to the fact that keyword matching with only titles and claims (as described in Sec. \ref{sec:the_system}) is intuitively more related to human online search behavior and can narrow the search space down with very high accuracy, whereas filtering document using term-based method e.g., TF-IDF directly on the entire document collection tends to impose more errors.
However, keyword matching does not maintain its performance on the difficult subset and suffers a 25 points drop on the oracle score because the method is essentially semantics-agnostic, i.e. can not reason by taking linguistic context into consideration.
This imposed difficulty is better handled by reranking based on Pageview frequency, TF-IDF, and the dNSMN, with the last one outperforming the other two on all the metrics. Though dNSMN and Pageview frequency ranking obtain comparable results on oracle score, the two methods are inherently different in that the former approaches document selection via advanced self-learned deep representations while the latter via demographic bias. Thus, we also experiments on combining the two methods by first re-ranking using Pageview and then dNSMN.
Finally, though the performances of all the methods are affected by increasing the number of retrieved documents from 5 to 10, the methods that use dNSMN merely suffered a 1 point drop on the retrieval accuracy on the entire dev set,
indicating that it's relatively more robust than other methods.

\paragraph{Sentence Selection Results}
Similar to the document retrieval setup, we evaluate the sentence selection performance on both the entire dev set and a difficult subset. The difficult subset for sentence selection is built by selecting examples in which the number of word-overlap between the claim and the ground truth evidence is below 2 and thus requires higher semantic understanding.
Neural networks with better lexical representations are intuitively more robust at selecting semantically related sentences than term weighting based methods. This fact is reflected in Table \ref{tab:sent_selection_ablation}, where although TF-IDF and the Max-pool Sentence Encoder obtain similar oracle FEVER scores (83.77\% and 84.08\%) and evidence recall (75.65\% and 76.13\%), the latter could achieve a much higher score for all metrics on the difficult subset. Note that for the entire dev set, the oracle score of the normally-trained (without annealed sampling) sNSMN (86.65\%) is higher than that of the Max-Pool sentence encoder (84.08\%) but on the difficult set, the sNSMN obtains a lower recall (68.34\%) compared to Max-Pool sentence encoder (73.68\%). This is due to the fact that the model with a stronger alignment mechanism will be more strict about selecting evidence and thus tends to trade accuracy for recall. This motivates our usage of annealed sampling in order to improve evidence recall. Although the annealed sampling reduces the evidence F1, we will explain later that this improvement of recall is important for the final FEVER Score.

\paragraph{Claim Verification Results}
\begin{table}[t]
\centering
\begin{tabular}{lccc}
\toprule
\multirow{2}{*}{\textbf{Model}} & \multirow{2}{*}{\bf{FEVER}} & \multirow{2}{*}{LA} & F1 \\
\cmidrule(lr){4-4}
& & & S/R/NEI
\\
\midrule
\textbf{Final Model} & \bf{66.14} & \bf{69.60} & \textbf{75.7}/69.4/\textbf{63.3}	\\
w/o WN and Num & 65.37 & 68.97 & 74.7/68.0/63.3  \\
w/o SRS (sent) & 64.90 & 69.07 & 74.5/\textbf{70.7}/60.7 \\
w. SRS (doc) & 66.05 & 69.69 & 75.6/70.0/62.8\\
Vanilla ESIM & 65.07 & 68.63 & 73.9/68.1/63.0
\\
\multicolumn{4}{l}{\footnotesize{\textit{\textbf{Data from sNSMN}}}}\\
\midrule
\textbf{Final Model} & 62.48 & 67.23 & 72.6/70.4/56.3 \\
\multicolumn{4}{l}{\footnotesize{\textit{\textbf{Data from TF-IDF}}}}\\
\bottomrule
\end{tabular}
\caption{Ablation study for verification (vNSMN). `WN'=WordNet feature, `Num'=number embedding, `Final Model'=vNSMN with semantic relatedness score feature only from sentence selection.
`SRS (sent)', `SRS (doc)' = Semantic Relatedness Score from document retrieval and sentence selection modules. {\bf FEVER} column shows strict FEVER score and LA column shows label accuracy without considering evidence. The last column shows F1 score of three labels. All models above line are trained with sentences selected from sNSMN for non-verifiable examples, while  model below is from TF-IDF.}
\label{tab:nli_ablation}
\end{table}

\begin{table}[t]
\centering
\begin{tabular}{cccccc}
\toprule
Threshold & \textbf{FEVER} & LA & Acc. & Recall & F1 \\
\midrule
0.5 & 66.15 & 69.64 & 36.50 & 86.69	& 51.37 \\
0.3 & 66.42 & 69.76 & 33.17 & 86.90	& 48.01  \\
0.1 & 66.43 & 69.67 & 29.83 & 86.97	& 44.42  \\
0.05 & \bf{66.49} & \bf{69.72} & 28.64 & 87.00 & 43.10  \\
\bottomrule
\end{tabular}
\caption{Dev set results (before evidence enhancement) for a vNSMN verifier making inference on data with different degrees of noise, by filtering with different score thresholds.}
\label{tab:nli_noise_tolerance_ablation}
\end{table}

\begin{table}[t]
\centering
\begin{tabular}{rc}
\toprule
Combination & \footnotesize{FEVER} \\
\midrule
\footnotesize{Pageview + dNSMN + sNSMN + vNSMN} & \textbf{66.59} \\
\footnotesize{dNSMN + sNSMN + vNSMN} & 66.50 \\
\footnotesize{Pageview + sNSMN + vNSMN} & 66.43 \\
\bottomrule
\end{tabular}
\caption{Performance of different combinations on dev set.}
\label{tab:combination_evaluation}
\end{table}

\begin{table}[t]
\centering
\begin{tabular}{lccc}
\toprule
Model & F1 & LA & \footnotesize{FEVER} \\
\midrule
\footnotesize{UNC-NLP} \scriptsize{(our shared task model)} & 52.96 & 68.21 & 64.21 \\
\footnotesize{UCL Machine Reading Group} &	34.97 & 67.62	& 62.52 \\
\footnotesize{Athene UKP TU Darmstadt} & 36.97	& 65.46 &	61.58 \\
\midrule
\footnotesize{UNC-NLP} \scriptsize{(our final model)} & 52.81	& 68.16 &	\textbf{64.23} \\
\bottomrule
\end{tabular}
\caption{Performance of systems on blind test results.}
\label{tab:competition_results}
\end{table}

We also conduct ablation experiments for the vNSMN with the best retrieved evidence\footnote{The best retrieved evidence is extracted from our dNSMN and sNSMN models (trained with annealed sampling).} on the FEVER dev set. Specifically, we choose the vNSMN with semantic relatedness score feature only from sentence selection as our \textbf{Final Model} (because it obtains the best results on FEVER score), and make modifications based on that model for analyzing different add-ons. The results are included in Table \ref{tab:nli_ablation}. First of all, we see that WordNet features (WN) and number embedding (Num) is able to increase the FEVER score, specifically by improving roughly 1 point of F1 scores on both the ``\textsc{Supports}'' (from 74.7 to 75.7) and the ``\textsc{Refutes}'' (from 68.0 to 69.4) examples because ontological features from WordNet (e.g., symptoms, antonyms, and hypernymms) and ordinal numeral features provide discriminative and fine-grained relational information which is extremely useful for revealing entailment and contradiction relations. More importantly, by incorporating the semantic relatedness score from the sNSMN model into the downstream vNSMN model, we also observe a 1 point improvement on FEVER score and almost 3 points improvement on F1 score for ``\textsc{Not Enough Info}'' examples. This approach can be viewed as combining evidence extraction with verification, by providing the verifier the degree of trustworthiness for each evidence and helping it recognize subtle neural relations between evidence and the claim.
We also see that the vNSMN with semantic relatedness score from both document retrieval and sentence selection modules achieves comparable (slightly worse) results to the vNSMN with semantic relatedness score from only the sentence selection module (hence, we use the latter for our final model). Our intuition of this phenomenon is that the document extraction subtask is two hops away from the claim verification subtask and hence its annotation supervision is less useful than the sentence selection subtask which is only one hop away. We also compare our vNSMN (66.14\% on FEVER) with vanilla ESIM model (65.07\% on FEVER) and the results on all metrics demonstrate that our architecture is better at modeling semantic matching. Lastly, we compare the performances of the same vNSMN with different training data for non-verifiable examples. The change of training data induces significant drops on both FEVER accuracy and F1 for the ``Not Enough Info'' example, highlighting the importance of the quality of upstream training data for neural inference model.

\paragraph{Noise Tolerance of vNSMN}

We evaluate the robustness of the vNSMN to noisy evidence during inference by setting different probability thresholds for filtering upstream evidence where the default 2-way softmax classification threshold is $0.5$. By reducing this value, we are allowing less confident evidence to be selected for downstream vNSMN. In Table~\ref{tab:nli_noise_tolerance_ablation}, we can see that the overall FEVER score is slightly increasing with the decrease of the threshold, indicating that the vNSMN is immune to noise. The findings encourage our usage of annealed sampling during sentence selection training and providing high recall evidence for the final fact verification model. We set threshold to $0.05$ for sentence.

\paragraph{Combination Evaluation and Final Result}
Since the KM + dNSMN and KM + Pageview + dNSMN setups get similar results on document retrieval (see Table \ref{tab:doc_retrieval_ablation}), we also compare their final FEVER results using the best downstream model (see Table \ref{tab:combination_evaluation}). Based on these dev FEVER results, we choose our final model as the combination of Pageview and dNSMN for blind test evaluation (though the non-Pageview neural-only model is still comparable). 
Finally, in Table \ref{tab:competition_results}, we present blind test results of our final system together with the top 3 results on the FEVER Shared Task leaderboard\footnote{\url{http://fever.ai/task.html}}. Our final system (Pageview + dNSMN + sNSMN + vNSMN) is able to get comparable results with our earlier shared task rank-1 system (Pageview + sNSMN + vNSMN), achieving the new state-of-the-art on FEVER.

\section{Conclusion}    
We addressed the fact verification FEVER task via a three-stage setup of document retrieval, sentence selection, and claim verification. We develop consistent and joint neural semantic matching networks for all three subtasks, along with Pageview, WordNet, and inter-module features, achieving the state-of-the-art on the task.

\section*{Acknowledgments}
We thank the reviewers for their helpful comments, and support via Verisk, Google, and Facebook research awards.

\bibliography{ref}
\bibliographystyle{aaai}

\appendix
\section{Supplemental Material}
\label{sec:supplemental}

\subsection{Dataset Splits and Scoring}
FEVER consists of 185,455 annotated claims, together with 5,416,537 Wikipedia documents containing roughly 25 million sentences as potential evidence. In our experiment, we use the same dataset partition as in the FEVER Shared Task, in which labels in the development (dev) set are intentionally balanced. Table \ref{tab:fever_data_split} shows the statistics of the dataset.
\begin{table}[ht]
\centering
\begin{tabular}{lccc}
\toprule
\textbf{Split} & \textsc{Supports} & \textsc{Refutes} & \textsc{NEI} \\ 
\midrule 
Training & 80,035 & 29,775 & 35,639 \\
Dev & 6,666 & 6,666 & 6,666 \\
Test & 6,666 & 6,666 & 6,666 \\
\bottomrule
\end{tabular}

\caption{Dataset split sizes for \textsc{Supports}, \textsc{Refutes} and \textsc{NotEnoughInfo} (\textsc{NEI}) classes.}
\label{tab:fever_data_split}
\end{table}

As described above, the single prediction is considered to be correct if and only if both the label is correct and the predicted evidence set (containing at most five sentences\footnote{This constraint is imposed in the FEVER Shared Task because in the blind test set all claims can be sufficiently verified with at most 5 sentences of evidence.}) covers the annotated evidence set. This score is named as \textbf{FEVER Score}. For insightful analyses and diagnoses of each sub-module in the system, we also evaluate the performance of different sub-modules on other metrics (F1, Precision, Recall, and Accuracy).
\subsection{Keyword Matching}
The keyword matching process in the document retrieval module consists of the following three rules:
\begin{itemize}
    \item \textbf{Exact Matching}: The document will be selected if there is an exact matching between the title and a span of text in the input claim. The matching is case insensitive except for the first letter (e.g., document ``YouTube'' would match ``YouTube'', ``Youtube'' but not ``youtube''). For document titles with disambiguation information in the parentheses, such as ``Savages (band)'', the matching will be triggered without considering the text in the parentheses (e.g., for claim ``Savages was exclusively a German film'', the processor will return ``Savages'', ``Savages (band)'', ``Savages (2012 film)'', etc.). 
    \item \textbf{First Article Elimination.} If the claim starts with `a', `an' or `the', we remove it and conduct the same exact matching rule on the new claim.
    \item \textbf{Singularization}: If no document title is returned with the above matching rule, we singularize every token in the claim and reapply exact matching rules on the new claim.
\end{itemize}

\begin{figure}[ht]
\begin{small}
\fbox{
\begin{minipage}{\linewidth}
\textbf{Claim:} \ul{Nicholas Brody} is a character on \ul{Homeland}.\\
\textbf{Retrieved Evidence:}\\
\textit{[{\ttfamily wiki/Homeland}]}\\
Homeland is the first novel in The Dark Elf Trilogy, a prequel to The Icewind Dale Trilogy, written by R. A. Salvatore and follows the story of Drizzt Do'Urden from the time and circumstances of his birth and his upbringing amongst the drow (dark elves).\\

\textit{[{\ttfamily wiki/Nicholas\_Brody]}}\\
GySgt. Nicholas "Nick" Brody, played by actor Damian Lewis , is a fictional character on the American television series Homeland on Showtime, created by Alex Gansa and Howard Gordon.\\

\textbf{Label:} Support
\end{minipage}}

\caption{An example of positive output. The model is able to correctly identify the evidence and verify the claim.}
\label{fig:pos_example}
\end{small}
\end{figure}

\begin{figure}[ht]
\begin{small}
\fbox{
\begin{minipage}{\linewidth}
\textbf{Claim:} \ul{Munich} is the capital of \ul{Germany}.\\
\textbf{Retrieved Evidence:}\\
\textit{[{\ttfamily wiki/Germany}]}\\
Germany's capital and largest metropolis is Berlin, while its largest conurbation is the Ruhr (main centres: Dortmund and Essen).\\

\textit{[{\ttfamily wiki/Munich]}}\\
Munich is the capital and largest city of the German state of Bavaria, on the banks of River Isar north of the Bavarian Alps.\\
Following a final reunification of the Wittelsbachian Duchy of Bavaria, previously divided and sub-divided for more than 200 years, the town became the country's sole capital in 1506.\\
Having evolved from a duchy's capital into that of an electorate (1623), and later a sovereign kingdom (1806), Munich has been a major European centre of arts, architecture, culture and science since the early 19th century, heavily sponsored by the Bavarian monarchs.\\

\textbf{Label:} Support
\end{minipage}}

\caption{An example of negative output. The model is able to correctly select the evidence but fail to reveal the correct logical relation between the evidence and the claim.}
\label{fig:neg_example}
\end{small}
\end{figure}

\subsection{Evidence Enhancement}
\label{sec:sup:evidence_enhancement}
Wikipedia document contains hyperlinks that are mapped to other related documents. Therefore, we consider a two-hop evidence selection that selects new evidential sentences from an existing evidence set by adding hyperlinked documents of the existing set. The process is described below:
\begin{itemize}
    \item We iterate over all the input sentences in the existing evidence set, we add all the documents that are hyperlinked by these sentences into a candidate document set.
    \item We conduct the same evidence retrieval process using the dNSMN and the sNSMN described in the paper from the candidate set and output the resulting new evidence set (enhancement evidence set).
    \item We select one sentence with the highest semantic relatedness score produced by the sNSMN from enhancement evidence set and add it back to the existing evidence set.
\end{itemize}

\paragraph{Evidence Enhancement Results}
For our final model, the evidence enhancement can improve 0.2\% OFEVER for evidence retrieval and 0.08\% on the final FEVER scores.

\subsection{Example outputs}

Fig. \ref{fig:pos_example} shows a positive example output demonstrating the procedure of the system. Fig. \ref{fig:neg_example} contains a negative example that requires world knowledge for reasoning. Due to representation limitation, the model fails to capture the exclusiveness nature of capitalizing relation and is confused by the existence of lexical clue ("capital") in the evidence, resulting in an incorrect prediction. This promotes future work to integrate more world knowledge into the model for natural language inference.

\end{document}